%% file: TKDD.tex
\documentclass[format=acmsmall, review=false, screen=true]{acmart}

\AtBeginDocument{%
  \providecommand\BibTeX{{%
    \normalfont B\kern-0.5em{\scshape i\kern-0.25em b}\kern-0.8em\TeX}}}

\setcopyright{acmcopyright}
\copyrightyear{2021}
\acmYear{2021}
\acmDOI{10.1145/1122445.1122456}

\acmJournal{TKDD}
\acmVolume{0}
\acmNumber{0}
\acmArticle{111}
\acmMonth{1}

\usepackage{pifont}
\usepackage{subfigure}
\usepackage{multirow}
\usepackage{amsmath}

\newcommand{\ie}{\textit{i}.\textit{e}.}
\newcommand{\eg}{\textit{e}.\textit{g}.}
\newcommand{\etal}{\textit{et} \textit{al}.}

\begin{document}

\title{Deep Graph Matching and Searching for Semantic Code Retrieval}


\author{Xiang Ling}
\authornote{Xiang Ling and Lingfei Wu contribute equally to this research.}
\orcid{0000-0002-7377-7844}
\affiliation{%
  \institution{Zhejiang University}
  \country{China}
}
\email{lingxiang@zju.edu.cn}

\author{Lingfei Wu}
\authornotemark[1]
\affiliation{%
  \institution{IBM T. J. Watson Research Center}
  \country{USA}
}
\email{lwu@email.wm.edu}

\author{Saizhuo Wang}
\affiliation{%
  \institution{Zhejiang University}
  \country{China}
}
\email{szwang@zju.edu.cn}

\author{Gaoning Pan}
\affiliation{%
  \institution{Zhejiang University}
  \country{China}
}
\email{szwang@zju.edu.cn}

\author{Tengfei Ma}
\affiliation{%
  \institution{IBM T. J. Watson Research Center}
  \country{USA}
}
\email{Tengfei.Ma1@ibm.com}

\author{Fangli Xu}
\affiliation{%
  \institution{Squirrel AI Learning}
  \country{USA}
}
\email{lili@yixue.us}

\author{Alex X. Liu}
\affiliation{%
  \institution{Ant Group}
  \country{China}
}
\email{alexliu@antfin.com}

\author{Chunming Wu}
\authornote{Chunming Wu and Shouling Ji are the co-corresponding authors.}
\affiliation{%
  \institution{Zhejiang University}
  \country{China}
}
\affiliation{%
  \institution{Zhejiang Lab}
  \country{China}
}
\email{wuchunming@zju.edu.cn}

\author{Shouling Ji}
\authornotemark[2]
\affiliation{%
  \institution{Zhejiang University}
  \country{China}
}
\email{sji@zju.edu.cn}


\renewcommand{\shortauthors}{Ling and Wu, et al.}

\begin{abstract}
Code retrieval is to find the code snippet from a large corpus of source code repositories that highly matches the query of natural language description.
Recent work mainly uses natural language processing techniques to process both query texts (\ie, human natural language) and code snippets (\ie, machine programming language), however neglecting the deep structured features of query texts and source codes, both of which contain rich semantic information.
In this paper, we propose an end-to-end deep graph matching and searching (DGMS) model based on graph neural networks for the task of semantic code retrieval.
To this end, we first represent both natural language query texts and programming language code snippets with the unified graph-structured data, and then use the proposed graph matching and searching model to retrieve the best matching code snippet.
In particular, DGMS not only captures more structural information for individual query texts or code snippets but also learns the fine-grained similarity between them by cross-attention based semantic matching operations.
We evaluate the proposed DGMS model on two public code retrieval datasets with two representative programming languages (\ie, Java and Python).
Experiment results demonstrate that DGMS significantly outperforms state-of-the-art baseline models by a large margin on both datasets.
Moreover, our extensive ablation studies systematically investigate and illustrate the impact of each part of DGMS.
\end{abstract}

\ccsdesc[500]{Information systems~Retrieval models and ranking}
\ccsdesc[300]{Information systems~Multimedia and multimodal retrieval}
\ccsdesc[100]{Computing methodologies~Natural language processing}
\ccsdesc[100]{Computing methodologies~Neural networks}

\keywords{Neural networks, graph representation, source code retrieval}

\thanks{This work was partly supported by
the National Key R\&D Program of China Under No. 2018YFB0804102, 2020YFB1804705, and 2020YFB2103802;
NSFC under No. 61772466, U1936215, and U1836202;
the Key R\&D Program of Zhejiang Province under No. 2020C01021, 2020C01077, and 2021C01036; 
the Major Scientific Project of Zhejiang Lab under No. 2018FD0ZX01;
the Zhejiang Provincial Natural Science Foundation for Distinguished Young Scholars under No. LR19F020003; 
and the Fundamental Research Funds for the Central Universities (Zhejiang University NGICS Platform).
}

\maketitle

\input{1Introduction}
\input{2.1Method}
\input{2.2Method}
\input{3Experiment}
\input{4RelatedWork}
\input{5Conclusion}



\bibliographystyle{ACM-Reference-Format}
\bibliography{6Reference.bib}


\end{document}

%% file: 1Introduction.tex
\section{Introduction}
With the advent of massive source code repositories such as GitHub\footnote{\url{https://www.github.com/}}, Bitbucket\footnote{\url{https://www.bitbucket.org/}}, and GitLab\footnote{\url{https://about.gitlab.com/}}, 
code retrieval over billions of lines of source codes has become one of the key challenges in software engineering~\cite{allamanis2018survey}.
Given a natural language query (\ie, human natural language) that describes the intent of program developers, the goal of code retrieval is to find the best matching code snippet (\ie, machine programming language) from a large corpus of code snippets in source code repositories.
Code retrieval tools can not only help program developers to find standard syntax or semantic usages of a specific programming language or library, but also help them to quickly retrieve previously written code snippets for certain functionality and then reuse those written code snippets, which largely accelerate software development for program developers.

To deal with the task of code retrieval, traditional approaches~\cite{sindhgatta2006using,linstead2009sourcerer,mcmillan2011portfolio,hill2011improving, lv2015codehow} mainly employ information retrieval techniques that treat source codes as a collection of documents and perform keyword searching over them.
For example,
{Sindhgatta}~\cite{sindhgatta2006using} developed a source code search tool - JSearch, which employs text based information retrieval techniques and indexes source code of Java programming languages by extracting features of syntactic entities.
{Linstead}~\etal~\cite{linstead2009sourcerer} proposed sourcerer, an information retrieval based tool that incorporates both textual information and structural fingerprints (\ie, code-specific heuristics) of source code and exploits the PageRank~\cite{pagerank} algorithm for ranking in code retrieval.
{Lv}~\etal~\cite{lv2015codehow} implemented CodeHow that applies the Extended Boolean model to combine the impact of both text similarity and potential APIs expanding on code retrieval.
However, all those information retrieval-based approaches have difficulty in understanding the semantics of both query texts and source codes.

More recently, advanced deep learning techniques, especially natural language processing methods, have been applied to learn deeper semantic information of both query and source code for code retrieval tasks.
One representative approach is DeepCS~\cite{gu2018deep} that applies deep neural networks to solve code retrieval tasks for the first time in 2018.
DeepCS first uses Recurrent Neural Network (RNN)~\cite{cho2014learning} or multi-layer perceptron (MLP)~\cite{collobert2004links} to encode both code snippets and query texts into a vector representation respectively and then compute the similarity score between their representations.
Other follow-up work~\cite{gu2018deep,sachdev2018retrieval,cambronero2019deep,yao2019coacor,wan2019multi,husain2019codesearchnet,haldar2020aMulti} is similar to DeepCS with only a slight difference in determining the encoding models.
Conceptually, these approaches usually apply two individual sequence encoder models (\eg, MLP, 1D-CNN~\cite{kim2014convolutional}, LSTM~\cite{hochreiter1997long}, or even BERT~\cite{devlin2018bert} encoder) for both the query text and the code snippets, and then rank these code snippets according to the similarity score between the learned distributed representations of the query text and every code snippet in the candidate set.
However, we argue that these approaches still suffer from two major challenges:
1) sequence encoding models cannot capture the structural information behind, especially for the source codes in which various dependency features include long-range dependencies may exist (\eg, the same identifiers may be operated in many places of source code);
2) lack of exploration of the semantic relationship between query texts and code snippets makes these models unable to align their distributed representation with fine-granularity.

To address these aforementioned challenges, in this paper, we propose a novel \textbf{D}eep \textbf{G}raph \textbf{M}atching and \textbf{S}earching (\textbf{DGMS}) model for representation learning and matching of both query texts and source codes.\footnote{For brevity, we interchangeably use ``source code'', ``code'' and ``code snippet'' for the source code, as well as ``query text'', ``query'' and ``text'' for the natural language description of the query.}
Our proposed DGMS model is based on two key insights.
Firstly, since higher abstraction requires capturing more semantic information, if the representation learning of both query texts and source codes has a higher abstraction, it would be more powerful to learn their semantic information.
Intuitively, graph-structured data represents a much higher abstraction than plain sequences or tokens, and hence the proposed DGMS model use graph-structured data as the basis of the representation learning for code retrieval tasks.
More importantly, we implement a novel graph generation method that represents both query texts and source codes into a unified graph-structured data in which both structural and semantic information could be largely retained.
Secondly, instead of simply encoding each graph-structured data as a graph-level embedding vector and calculating the cosine similarity between them, we first leverage the power of graph neural networks (GNN) to learn all node embeddings for graphs to capture semantic information for individual text graph or code graph.
Then, we propose a semantic matching operation based on the cross-attention mechanism to explore more fine-grained semantic relations between the text graph and the corresponding code graph for updating the embedding of each node in both graphs.

Specifically, as depicted in Figure~\ref{fig:model}, DGMS consists of three key modules:
\ding{182}~\textit{Graph encoding} module extracts deep structural representation from both query texts and source codes.
It employs graph neural networks to encode both text and code graphs individually with each node encoded with contextual embedding.
\ding{183}~\textit{Semantic matching} module utilizes a cross-attention based matching operation to obtain the semantic intersection relationship between the text graph and corresponding code graph and update the embedding of each node.
\ding{184}~\textit{Code searching} is performed by first aggregating all node embeddings for both text and code graphs to obtain two graph-level embeddings and then computing their ultimate similarity score.
Finally, we leverage the ranking loss to train the model in an end-to-end fashion.

To demonstrate the effectiveness of the proposed DGMS model, we systematically investigate the performance of DGMS compared with 7 state-of-the-art baseline models on two public code retrieval datasets: \textit{FB-Java} and \textit{CSN-Python} that are built from two different and representative programming languages \ie, Java and Python, respectively.
The experimental results show that our model significantly outperforms baselines by a large margin on both datasets in terms of all evaluation metrics.
We also conduct ablation studies to evaluate the contribution of each part of our DGMS model.
Our code can be available at \url{https://github.com/kleincup/DGMS}.
To summarize, we highlight the main contributions of our work as follows:
\begin{itemize}
    \item To the best of our knowledge, it is the first approach using the unified graph-structured data to represent both query texts and code snippets whose structural and semantic information is largely preserved and learned.
    \item We propose DGMS, a deep graph matching and searching model for semantic representation learning of both query texts and code snippets. DGMS can not only capture semantic information for individual the query text or code snippet, but also explore fine-grained semantic relations between them.
    \item We conduct extensive experiments on two public code retrieval datasets from two representative programming languages (\ie, Java and Python) and our results demonstrate the superior performance of our model over the state-of-the-art baseline methods.
\end{itemize}

\textbf{Roadmap:} The rest of the paper is organized as follows.
We represent both query texts and source codes into a unified graph-structured data in Section~\ref{sec:graph}, and describe the proposed DGMS model for graph representation matching and searching in Section~\ref{sec:model}.
The performance of the proposed DGMS model is systematically evaluated and analyzed in Section~\ref{sec:experiments}.
Section~\ref{sec:related_work} surveys related work and Section~\ref{sec:conclusion} finally concludes this work.

%% file: 2.1Method.tex
\section{Semantic Code Retrieval: From the Unified Graph Perspective}\label{sec:graph}
Before introducing the proposed DGMS model architecture for semantic code retrieval tasks,
in this section, we first introduce how we generate graph-structured data for both query texts and code snippets (Section~\ref{sub_sec:graph_generation}) and then define the code retrieval task from the unified graph perspective (Section~\ref{sub_sec:task_definition}).

\subsection{Graph Generation}\label{sub_sec:graph_generation}
\begin{figure}
\centering
\subfigure[An example java function with its natural language description and the code snippet.]{\label{fig:vis:sub1}
\includegraphics[width=0.54\linewidth]{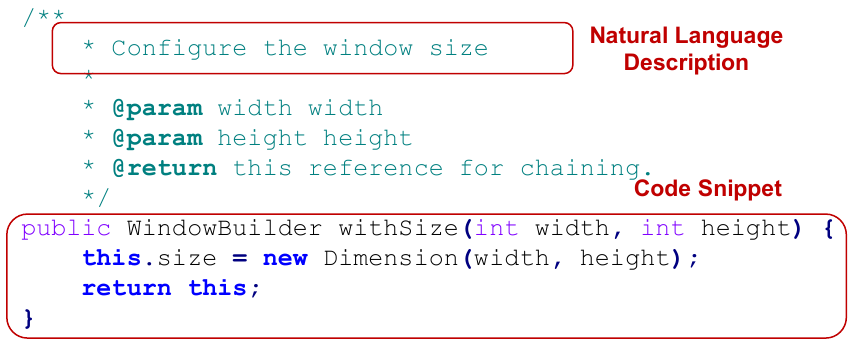}}
\hfill
\subfigure[The generated text graph of the natural language description in Figure~\ref{fig:vis:sub1}, in which there are 3 constituency symbols: S (the simple declarative clause), VP (verb phrase), and NP (noun phrase).]{\label{fig:vis:sub2}
\includegraphics[width=0.44\linewidth,]{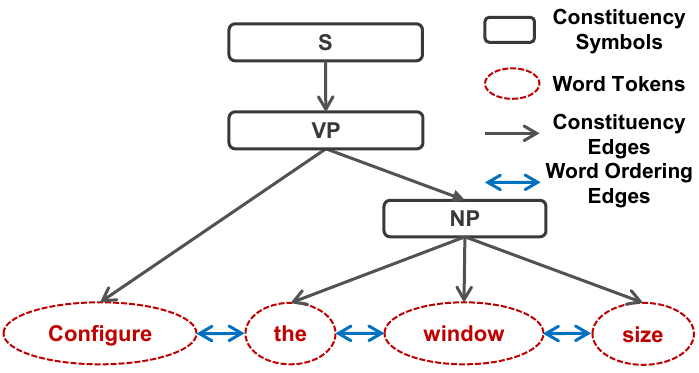}}
\vfill
\subfigure[The generated code graph of the java function in Figure~\ref{fig:vis:sub1}.]{\label{fig:vis:sub3}
\includegraphics[width=0.98\linewidth]{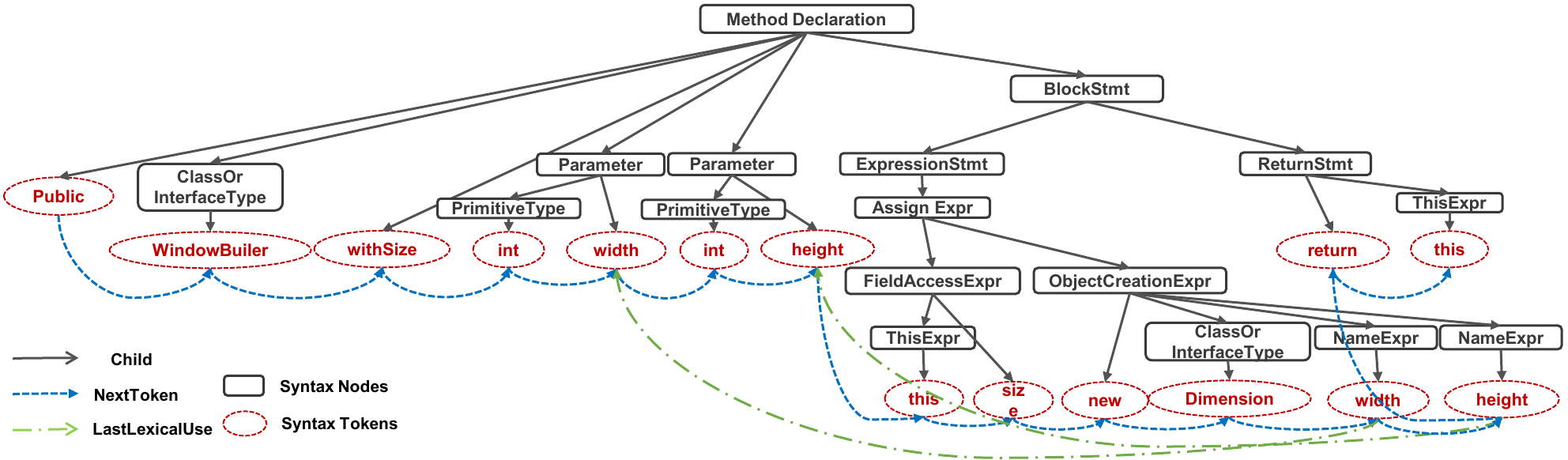}}
\caption{An example java function is shown in Figure~\ref{fig:vis:sub1} with its generated text graph in Figure~\ref{fig:vis:sub2} and the generated code graph in Figure~\ref{fig:vis:sub3}.
}
\label{fig:vis:total}
\end{figure}

Instead of simply treating both query texts and code snippets as plain tokens or sequences, we argue that both of them have rich important semantic structure information (\eg, various dependency features).
Thus, we propose to represent both the query text and code snippet with graph-structured data.
One real example\footnote{Source:~\url{https://github.com/onyxbits/raccoon4/blob/master/src/main/java/de/onyxbits/weave/swing/WindowBuilder.java\#L135} (last accessed on January 2020).} of query text and code snippet as well as their corresponding graph representations are shown in Figure~\ref{fig:vis:total}.
Below, we will introduce how we generate the text graph (Section~\ref{sub_sub_sec:text_graph}) and the code graph (Section~\ref{sub_sub_sec:code_graph}) from the corresponding query text and code snippet in detail.
\subsubsection{Generating Text Graphs}\label{sub_sub_sec:text_graph}
For the query text in the form of natural languages, we build the text graph based on the constituency parse tree~\cite{Jurafsky:2019} and word ordering features, which provide both constituent and ordering information of sentences.
We argue that the constituency features from the constituency parse tree actually represent the phrase structure of sentences and thus semantic information can be obtained.
In general, the most widely used formal system for modeling constituent structure in natural languages is context-free grammar~\cite{chomsky1956three}.
A context-free grammar consists of a lexicon of symbols (\eg, words) and a set of grammar rules.
Each of the rules expresses the ways that symbols of the language can be grouped and ordered together.

Specifically, the context-free grammar for natural languages can be formally defined as a four-tuple $\langle \mathbb{N}, \mathbb{T}, \mathbb{R}, s \rangle$.
$\mathbb{N}$ is a set of non-terminal symbols;
$\mathbb{T}$ is a set of terminal symbols that disjoint from $\mathbb{N}$;
$\mathbb{R}$ is a set of grammar rules $\mathbb{R} \colon \mathbb{N} \rightarrow (\mathbb{T}~\cup~\mathbb{N})^*$ that map a non-terminal symbol to a list of its children (a string of symbols from the infinite set of strings $(\mathbb{T}~\cup~\mathbb{N})^*$);
and $s \in \mathbb{N}$ is the designated root symbol.
Taking Figure~\ref{fig:vis:sub2} as an example, $\mathbb{N}$ is the set of \{S, VP, NP\} and $\mathbb{T}$ is the set of \{``Configure'', ``the'', ``window'', ``size''\}.
One grammar rule in $\mathbb{R}$ can be ``$\text{VP} \rightarrow \text{Verb}~~\text{NP}$'', expressing that one kind of verb phrase (VP) can be composed of a verb in English followed by a noun phrase (NP).
Our text graph generation is based on the constituency parse tree where nodes express the terminal or non-terminal symbols and edges express its grammar rules, \ie, constituency edges (with black color) in Figure~\ref{fig:vis:sub2}.

On the other hand, we also incorporate the word ordering information of sentences into the text graph.
Specifically, we link the words (\ie, terminal symbols in the constituency tree) of sentences in a chain, which could capture the forward and backward contextual information of sentences.
That is, we add this kind of word ordering edges (with blue color in Figure~\ref{fig:vis:sub2}) into the constituency tree.
In summary, by combining both constituent and word ordering information of sentences into graph-structured data, we can generate a more informative text graph representation as shown in Figure~\ref{fig:vis:sub2}.

\subsubsection{Generating Code Graphs}\label{sub_sub_sec:code_graph}
Inspired by recent advances in solving source code related tasks, like code completion, variable naming, and code generation~\cite{allamanis2018learning,brockschmidt2019generative,fernandes2019structure,cvitkovic2019open}, we generate the code graphs using the \textit{program graph} structure~\cite{allamanis2018learning}, which is mainly based on the abstract syntax tree (AST) representation of source codes~\cite{slonneger1995formal}.
In fact, the AST of source codes is in analogy to the constituency parse tree of natural language texts.
Both of them share a similar spirit for representing either source codes or query texts.
In particular, an AST for a code snippet is a tuple $\langle \mathbb{N}, \mathbb{T}, \mathbb{X}, \Delta, \phi, s \rangle$,
where $\mathbb{N}$ is a set of non-terminal nodes,
$\mathbb{T}$ is a set of terminal nodes,
$\mathbb{X}$ is a set of values,
$\Delta \colon \mathbb{N} \rightarrow (\mathbb{N} \cup \mathbb{T})^*$ is a function that maps a non-terminal node to a list of its children, 
$\phi \colon \mathbb{T} \rightarrow \mathbb{X}$ is a function that maps a terminal node to an associated value, 
and $s \in \mathbb{N}$ is the root node.

Specifically, a \textit{program graph} consists of syntax nodes (corresponding to terminal/non-terminal nodes in AST) and syntax tokens (values of terminal nodes in the original code).
\textit{Program graphs} use different kinds of edge types to model the syntactic and semantic relationship between nodes/tokens.
As shown in Figure~\ref{fig:vis:sub3}, the \textit{program graph} explores three edge types:
\textit{Child} edges are to connect syntax nodes in AST,
\textit{NextToken} edges are to connect each syntax token to its successor in the original code, 
\textit{LastLexicalUse} edges are to connect identifiers to their most recent lexical usage.
More introduction and implementation details can be found in~\cite{allamanis2018learning, fernandes2019structure}.

\textbf{Why we choose code graphs rather than other graphs like control-flow graphs?}
In short, the main reason we choose to represent the code snippet with the \textit{program graph} is that it shares a similar spirit to the text graph of the query texts, which allows us to represent both query texts and source codes from a unified graph perspective. 
On the other hand, the reason why we do not represent source code as a control-flow graph (CFG) is mainly because control-flow related graphs are too coarse-grained (\ie, each node is a sequence of instructions without jumps where one instruction consists of multiple lines of sequential tokens) to properly represent source codes for code retrieval tasks.

\subsection{Code Retrieval Task Definition: A Unified Graph Perspective}\label{sub_sec:task_definition}
In this subsection, we define the code retrieval task from a unified graph perspective.
Specifically, given a corpus of source codes $E$ with a total number of $|E|$ code snippets, the goal of code retrieval is to find the best matching code snippet $\hat{e}$ from $E$ according to the query text $q$ in the form of natural language.
Thus, we give the formulation of the definition of code retrieval task as follows.
\begin{equation}
    \hat{e} = \operatorname*{arg~max}_{e \in E} sim(q, e) = \operatorname*{arg~max}_{e \in E}sim(G_{q}, G_{e})
    \label{equation:definition} 
\end{equation}
The core of this code retrieval task is to compute the similarity score $sim(q, e)$ between the code snippet $q$ and query text $e$.
As described in above Section~\ref{sub_sec:graph_generation}, both code and text can be represented with graph-structured data using our graph generation methods.
Thus, we further formulate $sim(q, e)$ as $sim(G_{q}, G_{e})$, in which $G_{q}$, $G_{e}$ are graph representations for the text $q$ and code $e$, respectively.
In this paper, both $G_{q}$ and $G_{e}$ are represented as directed and labeled multi-graphs in which different edge types are encoded by labeled edges.
Specifically, the text graph $G_{q}$ is represented as $(\mathcal{V}_{q}, \mathcal{E}_{q}, \mathcal{R}_{q})$ with nodes $q_i \in \mathcal{V}_{q}$ and edges $(q_i, r, q_j) \in \mathcal{E}_{q}$, where $r \in \mathcal{R}_{q}$ denotes edge type.
Similarly, the code graph $G_{e}$ is represented as $(\mathcal{V}_{e}, \mathcal{E}_{e}, \mathcal{R}_{e})$.
The number of nodes of $G_{q}$ and $G_{e}$ is $M$ and $N$, respectively.
Furthermore, the important symbols and notations throughout the following sections of our paper are summarized in Table~\ref{tab:notation}.
\begin{table}
\centering
\caption{Important symbols and notations}
\label{tab:notation}
\begin{tabular}{cl}
    \toprule
    \textbf{Symbols}&   \textbf{Definitions or Descriptions}            \\
    \midrule
    $q$             &   query text                                      \\

    $e$             &   code snippet                                    \\

    $G_{q}$         &   graph representation of query text $q$          \\

    $G_{e}$         &   graph representation of code snippet $e$        \\

    $q_i$           &   $i$-th node in graph $G_{q}$                    \\

    $e_j$           &   $j$-th node in graph $G_{e}$                    \\

    $\mathbf{q}_i \in \mathbb{R}^{d}$  & node embedding vector of $q_i$ in graph $G_{q}$ with $d$ dimensions   \\

    $\mathbf{e}_j \in \mathbb{R}^{d}$  & node embedding vector of $e_j$ in graph $G_{e}$ with $d$ dimensions \\

    $\mathbf{H}_q \in \mathbb{R}^{d'}$  & graph-level embedding vector of graph $G_{q}$ with $d'$ dimensions  \\

    $\mathbf{H}_e \in \mathbb{R}^{d'}$  & graph-level embedding vector of graph $G_{e}$ with $d'$ dimensions \\

    $M$             & the number of nodes in graph $G_{q}$              \\

    $N$             & the number of nodes in graph $G_{e}$              \\
    \bottomrule
\end{tabular}
\end{table}

%% file: 2.2Method.tex
\section{Deep Graph Matching and Searching}\label{sec:model}
\begin{figure*}[t]
    \centering
    \includegraphics[width=0.99\linewidth]{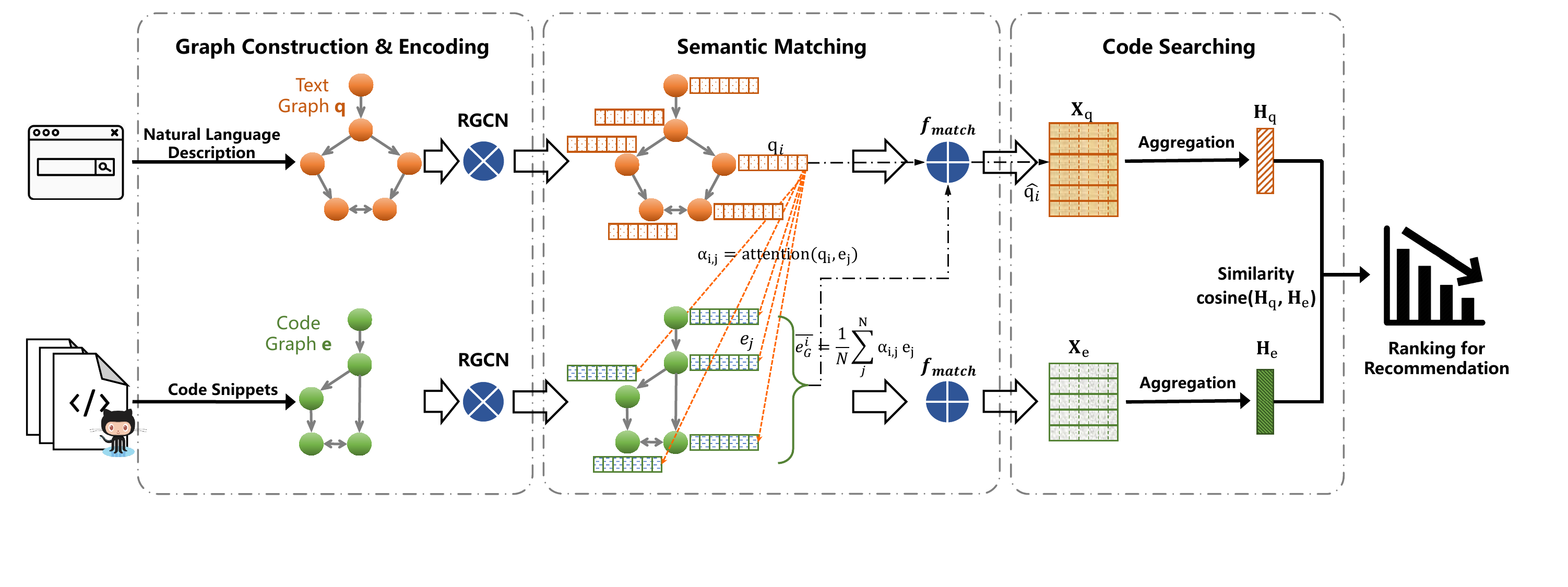}
    \caption{The general framework of deep graph matching and searching (DGMS) model.}
    \label{fig:model}
\end{figure*}

In this section, we introduce our proposed deep graph matching and searching (DGMS) model, which performs fine-grained text-code semantic matching and searching based on text graphs and code graphs.
Figure~\ref{fig:model} shows the overall DGMS architecture, consisting of three modules: graph generation, graph encoding, semantic matching, and code searching.
As graph generation has been introduced in previous Section~\ref{sec:graph}, we will detail the other three modules as follow.

\subsection{Graph Encoding}\label{sub_sec:graph_encoding}
In order to capture the structure and semantic information for individual text graphs and code graphs, in this graph encoding module, we first need to learn the node embeddings for each node in both text and code graphs.

Recently, the graph neural network (GNN) that adapts deep learning from image to graph-structured data has received unprecedented attention from both machine learning and data mining communities~\cite{wu2020comprehensive, rong2020deep}.
The main goal of GNN is to learn information representations (\eg, node or (sub)graph, etc) of graph-structured data in an end-to-end manner~\cite{bronstein2017geometric}.
Specifically, a GNN model takes a graph (\ie, its topological structures, node and/or edge attributes) as input, and output an embedding vector for each node of the input graph.
There is a large body of GNN models designed to learn node representations~\cite{scarselli2008graph, li2016gated, kipf2017semi, GraphSage:hamilton2017inductive, velivckovic2018graph, gilmer2017neural, xu2019powerful,chen2020iterative}.
With the learned node representation, various tasks on graphs can be performed such as link prediction~\cite{zhang2018link}, graph classification~\cite{wu2019scalable, Errica2020A}, and graph matching and similarity~\cite{bai2019simgnn, ling2020hierarchical,zhang2019kergm}, to name just a few.

Since both text graphs and code graphs are represented as directed and labeled multi-graphs (\ie, graphs with multiple labeled edges), we adopt one variant of GNNs -- Relational Graph Convolutional Networks (RGCNs)~\cite{schlichtkrull2018modeling} to learn their node embedding in this graph encoding module.
Other variants of graph neural networks can also be applied for the graph encoding as long as these variants could learn node embeddings of a graph with different edge types (see details in Section~\ref{sub_sub_sec:experiment_diff_graph_encoder}).

In particular, taking the text graph $G_{q}=(\mathcal{V}_{q}, \mathcal{E}_{q}, \mathcal{R}_{q})$ as an example, RGCN defines the propagation model for calculating the updated embedding vector $\mathbf{q}_{i}$ of each node $q_i \in \mathcal{V}_{q}$ in the text graph $G_{q}$ as follows,
\begin{equation}
    \mathbf{q}_{i}^{(l+1)} = \operatorname*{ReLU} \big(W_{\Theta}^{(l)} \mathbf{q}_{i}^{(l)} + \sum_{r \in \mathcal{R}_{q}} \sum_{j \in \mathcal{N}_{i}^{r}} \frac{1}{|\mathcal{N}_{i}^{r}|} W_{r}^{(l)} \mathbf{q}_{j}^{(l)}\big)
    \label{equation:RGCN}
\end{equation}
where $\mathbf{q}_{i}^{(l+1)}$ denotes the updated embedding vector of node $q_i$ in the $(l+1)$-th layer of RGCN,
$\mathcal{R}_{q}$ represents the set of relations (\ie, edge types),
$\mathcal{N}_{i}^{r}$ is the set of the neighbors of node $q_i$ under the edge type $r\in \mathcal{R}_{q}$,
$W_{\Theta}^{(l)}$ and $W_{r}^{(l)}$ are parameters of the RGCN model to be learned,
$\operatorname*{ReLU}$ denotes the activation function.

By encoding the graph-structured data for both the query text and code snippet with above RGCN model, we thus obtain both node embeddings $\mathbf{X}_{q} = \{\mathbf{q}_{i}\}_{i=1}^{M} \in \mathbb{R}^{(M,d)}$ for text graph $G_{q}$ and node embeddings $\mathbf{X}_{e} = \{\mathbf{e}_{j}\}_{j=1}^{N} \in \mathbb{R}^{(N,d)}$ for code graph $G_{e}$, in which $d$ represents the embedding dimensions of each node.  

\subsection{Semantic Matching}\label{sub_sec:semantic_matching}
After obtaining node embeddings with rich semantic information for each individual text graph or code graph, we propose a \textbf{cross-attention based semantic matching operation} $f_{\text{match}}$ for comparing and aligning every node embedding in one graph against the graph-level contextual embedding of the other graph.
The key idea of this semantic matching operation is to explore more fine-grained semantic relations between text graphs and the corresponding code graphs, and then update and enrich the node embeddings for both graphs.
We detail each step for semantic matching as follows.

\textbf{Attention:} we first compute the cosine attention similarity between all pairs of each node in one graph and all nodes in another graph.
Taking the node $q_i $ in $G_{q}$(\ie, $q_i \in \mathcal{V}_{q}$) as an example, we calculate a cross attention similarity between its node embedding $\mathbf{q}_i$ of node $q_i$ and the node embedding $\mathbf{e}_i$ of each node $e_j$ in graph $G_{e}$,
\begin{equation}
\alpha_{i,j} = \text{cosine}(\mathbf{q}_i, \mathbf{e}_j), \forall j = 1, \dots, N 
\end{equation}
where $\text{cosine}$ denotes the cosine similarity function.

\textbf{Context representation:} We then take the cosine attention (\ie, $\alpha_{i,j}$) as the weight of $\mathbf{e}_{j}$ and compute a weighted-average embedding over all node embeddings of $G_{e}$ by the corresponding cross attention scores with node $q_i$ in $G_{q}$, which yields a contextual global-level representation $\overline{\mathbf{e}_{G}^{i}}$ of $G_{e}$ with respect to node $q_i$ in $G_{q}$,
\begin{equation}
\overline{\mathbf{e}_{G}^{i}} = \frac{1}{N} \sum_{j}^{N} \alpha_{i,j} \mathbf{e}_{j}
\end{equation}

\textbf{Comparison:} Inspired by previous work of textual entailment in natural language processing~\cite{wang2017compare}, we consider two following matching operations $\operatorname*{Sub}$ and $\operatorname*{Mul}$, both of which transform $\mathbf{q}_i$ and $\overline{\mathbf{e}_{G}^{i}}$ into a vector $\hat{\mathbf{q}_i}$ to represent the comparison result.
We argue that the comparison result measures, to some extent, the alignment information between each pair of node embeddings in two graphs and thus can be used to update node embeddings for latter computing similarity between two graphs.
\begin{align}
    \hat{\mathbf{q}_i} &= \operatorname*{Sub}(\mathbf{q}_i, \overline{\mathbf{e}_{G}^{i}}) = (\mathbf{q}_i - \overline{\mathbf{e}_{G}^{i}}) \odot (\overline{\mathbf{e}_{G}^{i}} - \mathbf{q}_i) \label{equation:sub} \\
    \hat{\mathbf{q}_i} &= \operatorname*{Mul}(\mathbf{q}_i, \overline{\mathbf{e}_{G}^{i}}) = \mathbf{q}_i \odot \overline{\mathbf{e}_{G}^{i}} \label{equation:mul}
\end{align}
where $\odot$ denotes the element-wise multiplication operation;
the resulting $\hat{\mathbf{q}_i}$ is the updated node embedding vector and has the same embedding size as $\mathbf{q}_i$ or $\overline{\mathbf{e}_{G}^{i}}$.

Alternatively, the results of these two matching operations $\operatorname*{Sub}$ and $\operatorname*{Mul}$ can be concatenated to assemble another matching operation called $\operatorname*{SubMul}$, which is finally adopted by our final model DGMS,
\begin{align}
    \hat{\mathbf{q}_i} &= \operatorname*{SubMul} (\mathbf{q}_i, \overline{\mathbf{e}_{G}^{i}}) = \operatorname*{Concat} \big[ \operatorname*{Sub}(\mathbf{q}_i, \overline{\mathbf{e}_{G}^{i}}), \operatorname*{Mul}(\mathbf{q}_i, \overline{\mathbf{e}_{G}^{i}}) \big] \label{equation:submul}
\end{align}
where $\operatorname*{Concat}$ denotes the concatenation operation and the resulting $\hat{\mathbf{q}_i}$ is twice the node embedding size of $\mathbf{q}_i$ or $\overline{\mathbf{e}_{G}^{i}}$.

To sum up, after performing the above semantic matching operation $f_{match}$ for both text graph $G_{q}$ and code graph $G_{e}$,
we further update their node embeddings as $\mathbf{X}_{q} = \{\hat{\mathbf{q}_{i}}\}_{i=1}^{M} \in \mathbb{R}^{(M, d')}$ and $\mathbf{X}_{e} = \{\hat{\mathbf{e}_{j}}\}_{j=1}^{N} \in \mathbb{R}^{(N, d')}$, where $d'$ denotes the updated node embedding size (\ie, $d'= d$ for $\operatorname*{Sub/Mul}$, $d'= 2d$ for $\operatorname*{SubMul}$.).

\subsection{Code Searching}\label{sub_sec:code_search}
As shown in Equation~(\ref{equation:definition}), the core of code retrieval tasks is to calculate a similarity score between the code and query text, and thus to calculate the similarity between the two graph representations of both the query text and code snippet.

To perform code searching over graph representations of both codes and texts, we first aggregate the unordered node embeddings for each graph and obtain the corresponding graph-level representation vector for either the code graph or the text graph.
A simple and straightforward aggregation method is the max-pooling operation that calculates the maximum value for all node embeddings in one graph.
In this paper, we apply another pooling operation $\operatorname*{FCMax}$ on their node embeddings (\ie, $\mathbf{X}_{q}$ and $\mathbf{X}_{e}$).
Particularly, FCMax is a variant of the max-pooling operation following a fully-connected layer transformation.
\begin{align}
    \mathbf{H}_{q} &= \operatorname*{FCMax}(\mathbf{X}_{q}) = \operatorname*{max-pooling}(\operatorname*{FC}(\{\hat{\mathbf{q}_{i}}\}_{i=1}^{M}))\\
    \mathbf{H}_{e} &= \operatorname*{FCMax}(\mathbf{X}_{e}) = \operatorname*{max-pooling}(\operatorname*{FC}(\{\hat{\mathbf{e}_{i}}\}_{i=1}^{N}))
\end{align}
where $\operatorname*{FC}$ and $\operatorname*{max-pooling}$ denote the fully-connected layer and max-pooling operation.
It is evident that $\operatorname*{FCMax}$ offers a larger model capacity than max-pooling operation in theory, which is also supported by the latter experimental evaluation in Section~\ref{sub_sub_sec:experiment_diff_aggregation}.
The output dimension size of $\mathbf{H}_{q}$ depends on the hidden size of the fully-connected layer which is set the same as $d'$ (\ie, $\mathbf{H}_{q} \in \mathbb{R}^{d'}$ and $\mathbf{H}_{e} \in \mathbb{R}^{d'}$).

Next, to measure the similarity score between the query text and code snippet, \ie, $\operatorname{sim}(q, e)$ in Equation~(\ref{equation:definition}), 
we compute the cosine similarity between $\mathbf{H}_{q}$ and $\mathbf{H}_{e}$,
\begin{equation}
\operatorname{sim}(q, e) = \operatorname{sim}(G_q, G_e) = \operatorname*{cosine} (\mathbf{H}_{q}, \mathbf{H}_{e})
\label{equation:final_sim_cosine}
\end{equation}
According to Equation~(\ref{equation:definition}), we can perform code searching based on the similarity score between the two learned distributed representations.

\subsection{Model Training}\label{sub_sec:model_train}

In principle, our model can be trained in an end-to-end fashion on a large corpus of paired query texts and code snippets.
However, we use the document description of code snippet instead of the query text for model training, since there is no benchmark dataset that contains a large corpus of paired query texts and code snippets.
For example, the dataset in~\cite{li2019neural} only contains 287 Stack Overflow question/answer pairs and the dataset in~\cite{husain2019codesearchnet} contains 99 human-annotated query/code pairs per programming language.
Hence, it is impossible to train a code retrieval model based on those limited paired query texts and code snippets.

Specifically, each training sample in the training corpus $\mathbb{T}$ is a triple $\langle q, e, \ddot{e}\rangle$, which is constructed as follows:
for each code snippet $e$ and its corresponding documental text description $q$, we randomly select a negative sample code snippet $\ddot{e}$ from other code snippets in the corpus $\mathbb{T}$.
The goal of our model is to predict a higher cosine similarity $sim(q, e)$ than $sim(q, \ddot{e})$.
According to Equation~\ref{equation:final_sim_cosine}, we use the margin ranking loss~\cite{BMVC2016_119} for model optimization as follows,
\begin{align*}
  \mathcal{L}(\theta) &= \sum_{\langle q, e, \ddot{e}\rangle \in \mathbb{T}} \text{max}\big(0, \delta - \text{sim}(q, e) + \text{sim}(q, \ddot{e})\big) \\
  &= \sum_{\langle q, e, \ddot{e}\rangle \in \mathbb{T}} \text{max}\big(0, \delta - \text{cosine}(\mathbf{H}_{q}, \mathbf{H}_{e}) + \text{cosine}(\mathbf{H}_{q}, \mathbf{H}_{\ddot{e}})\big)
\end{align*}
where $\theta$ denotes the model parameters to be learned and $\delta$ is the margin parameter of margin ranking loss.

It is noted that our model is trained based on the siamese network~\cite{bromley1994signature} that uses the shared RGCN model to learn representations for both text graph and code graph.
The feature of sharing the parameters of the siamese network makes our model smaller, thus mitigating possible over-fitting and making the training process easier.

%% file: 3Experiment.tex
\section{Experiments}\label{sec:experiments}
In this section, we first introduce the benchmark datasets to be evaluated in Section~\ref{sub_sec:datasets}, describe the baseline models to be compared with our models in Section~\ref{sub_sec:baseline_models}, provide details of the experimental settings in Section~\ref{sub_sec:experimental_settings}, and finally present experimental results and discussion of the results in Section~\ref{sub_sec:experimental_results}.

\subsection{Datasets Description \& Preprocessing}\label{sub_sec:datasets}
We evaluate our DGMS model on two public code retrieval datasets: \textit{FB-Java} and \textit{CSN-Python} that are built from two different and representative programming languages, \ie, \textbf{Java} and \textbf{Python}.
The first \textit{FB-Java} dataset is built on the recently released benchmark dataset that the Facebook research team has been used to assess the performance of neural code search models~\cite{li2019neural,facebookresearch:releasedataset}.
As the released dataset provides the downloadable links of GitHub repositories and the original repositories might be deleted or removed by their owner for some reasons, we actually collect a total of 24,234 repositories with 4,679,758 functions or methods on the download date of October 11, 2019.
To make the dataset can be utilized to evaluate code retrieval tasks, we apply some preparatory operations as follows.
\begin{enumerate}
    \item For each downloaded method or function, we parse it to a pure code snippet and the corresponding docstrings (\ie, method-level documental description) if it has;
    \item We remove all methods without docstrings, as we treat the docstring as the query text in our evaluation;
    \item To make the dataset more realistic to evaluate the code retrieval tasks, we remove all methods whose pure code snippet contains less than 3 lines;
    \item We remove all methods whose docstring contains less than 3 words or contains non-English words;
    \item As some of those methods have duplicate docstrings (\eg, method overloading or overriding, etc.), we only keep one of the methods with duplicate docstrings.
\end{enumerate}
After that, we get a total of 249,072 pairs of source code and corresponding documental descriptions for the \textit{FB-Java} dataset.

To further show the effectiveness of our model on other programming languages, we also choose another dataset \textit{CSN-Python} from CodeSearchNet~\cite{husain2019codesearchnet}.
We apply similar preparatory operations on \textit{CSN-Python} as above, which essentially results in a total of 364,891 pairs of source code and documental description.

\begin{figure*}[htb]
\centering
\subfigure[FB-Java]{\includegraphics[width=0.45\linewidth]{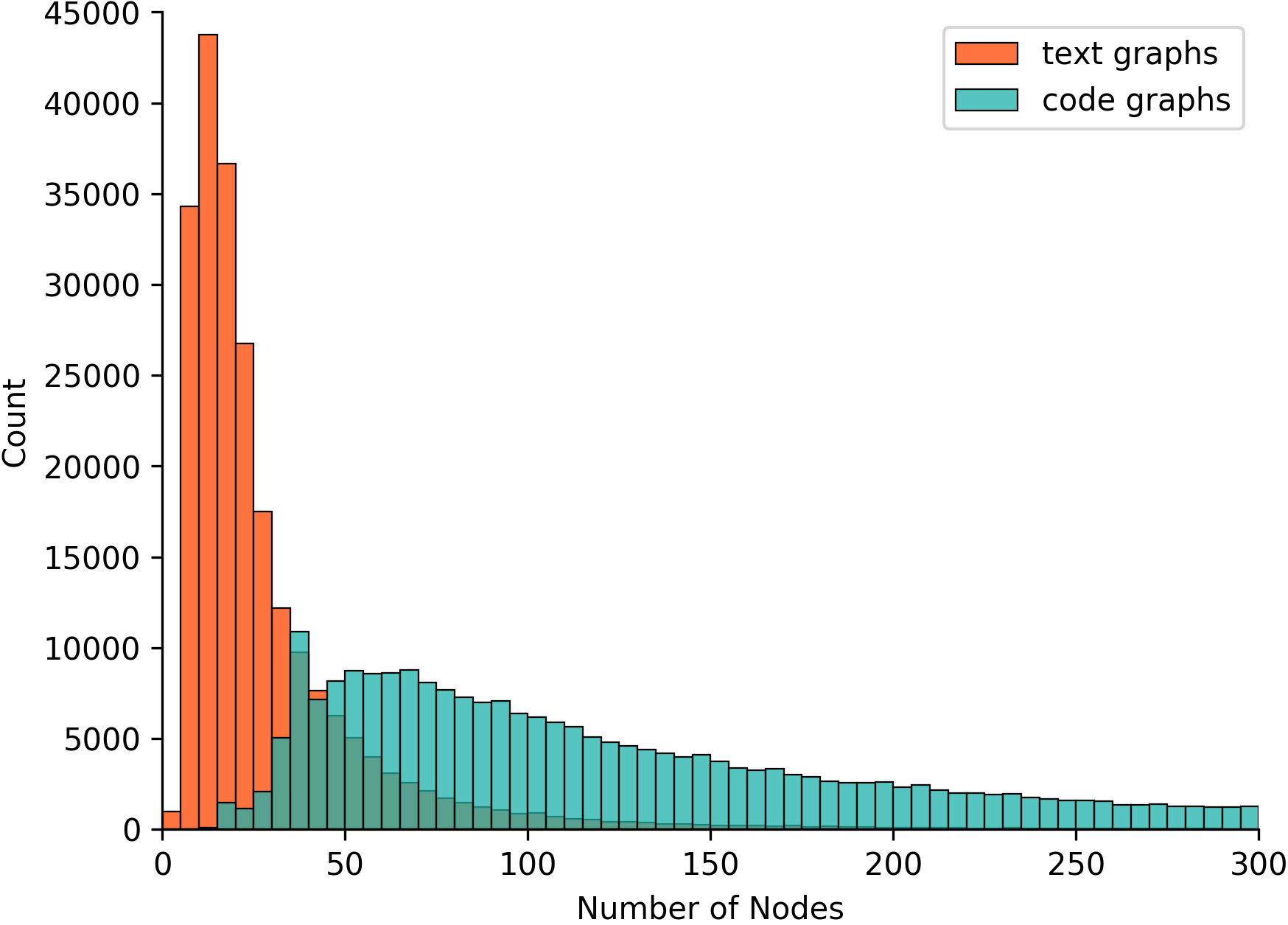}}
\subfigure[CSN-Python]{\includegraphics[width=0.45\linewidth]{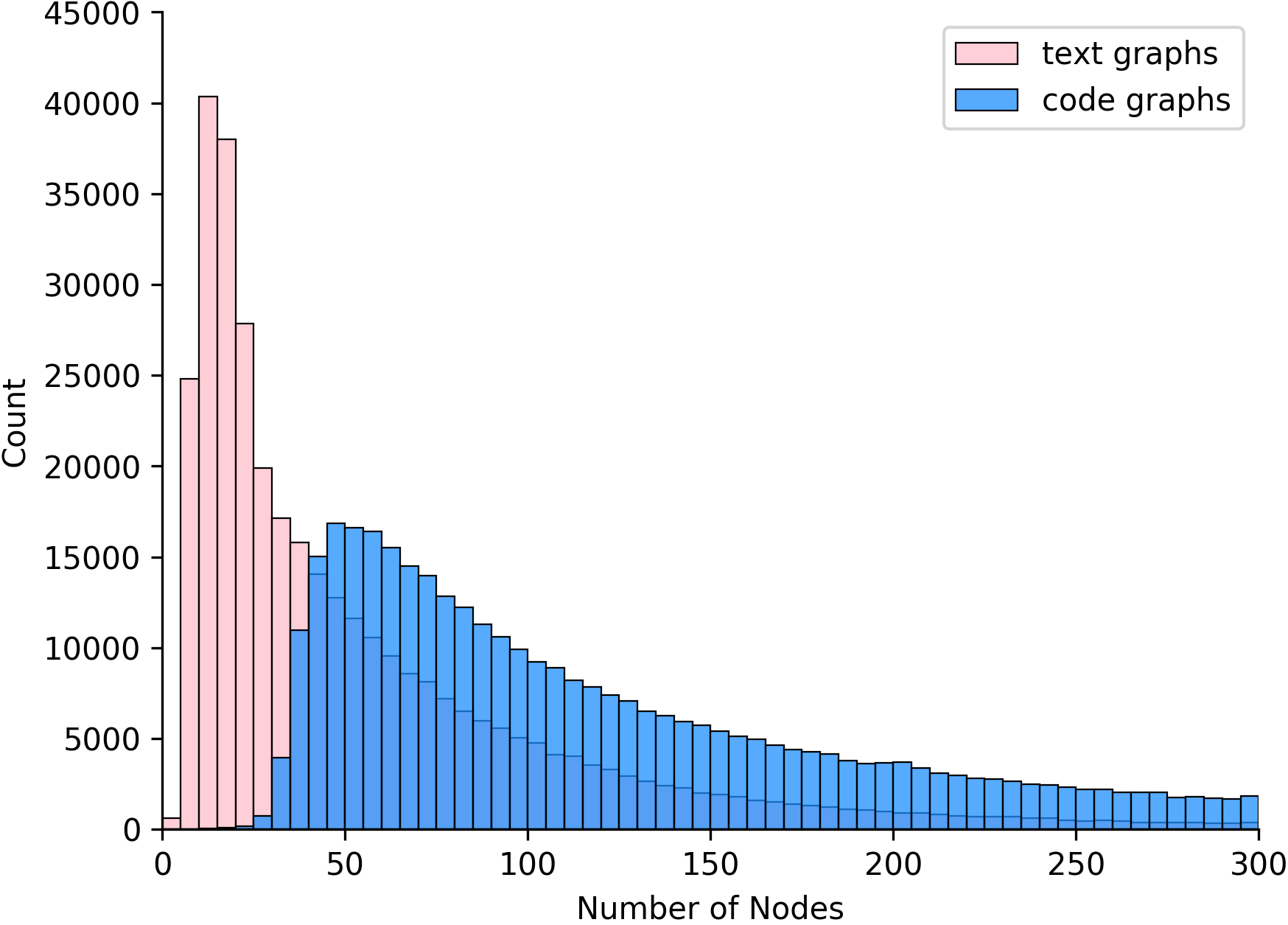}}
\caption{Graph size distribution for both datasets.}%
\label{fig:GraphDistribution}%
\end{figure*}

Since our model is built on the graph-structured data, we need to produce graph representations for both code snippets and text descriptions.
To produce the text graphs, we first generate the constituency tree of text description using the Stanford CoreNLP toolkit~\cite{manning-EtAl:2014:P14-5} and then link the terminal nodes in the constituency tree in a chain with bi-direction edges.
We build code graphs based on the open-sourced code in~\cite{fernandes2019structure}.
As illustrated in Figure~\ref{fig:GraphDistribution},
the number of nodes for both text graphs and code graphs in both datasets follows a typical long-tail distribution, we limit the number of nodes of all graphs to no more than 300, which keeps more than 90\% of the total dataset.
It is also obviously observed that the average number of nodes of text graphs is much less than that of code graphs for both datasets.
Finally, we split the dataset into training/validation/testing sets with statistics shown in Table~\ref{tab:data_statistics}.
\begin{table}[htb]
\centering
\caption{Datasets statistics.}
\begin{tabular}{lrrr}
\toprule
\textbf{Datasets} & \textbf{\# Training} & \textbf{\# Validation} & \textbf{\# Testing}  \\
\midrule
\textit{FB-Java}    &   216,259 &   9,000   &   1,000 \\
\textit{CSN-Python} &   312,189 &   17,215  &   1,000 \\
\bottomrule
\end{tabular}\label{tab:data_statistics}%
\end{table}

\subsection{Baseline Models}\label{sub_sec:baseline_models}
To evaluate the effectiveness of our model, we consider 7 state-of-the-art deep learning based models as baseline models for comparison.
Details of these baselines as well as their experimental settings are as follows.

\begin{itemize}

\item Neural Bag of Words (\textbf{Neural BoW}), \textbf{RNN}, \textbf{1D-CNN}, and \textbf{Self-Attention} are 4 baseline models provided by~\cite{husain2019codesearchnet} that first encode both the code and text with the corresponding neural network and then compute a similarity score between the representations of both code and text.
In particular, the open-sourced implementation of the RNN baseline employs LSTM as the encoder, and Self-Attention adopts the BERT encoder~\cite{devlin2018bert} in their implementation.
To avoid over-tuning, we use their original hyper-parameters and experimental settings.

\item \textbf{DeepCS}~\cite{gu2018deep} captures the semantic information of code from 3 perspectives: method name, API call sequence, and code tokens.
It first separately encodes the 3 different sequences with RNN or MLP,
and then fuse them to get code representation.
For the text, the authors simply encode its tokens with RNN. 
Finally, DeepCS computes the cosine similarity score between the representations of code snippets and texts. 
However, the extraction method of API call sequences is based on heuristic approaches that are specific to the Java language and do not work for other programming languages.
Thus, we only use method names and code tokens as the semantic features of source codes in our experiments.

\item \textbf{UNIF} maps both code and query tokens with learnable embeddings~\cite{cambronero2019deep}.
To generate their representations, all token embeddings of the code snippet are aggregated by attention weighted sum with learnable attention vectors, while query token embeddings are simply pooled on average.
The cosine distance is used as the similarity metric.
In our evaluations, we set the token embedding dimension to 100.

\item \textbf{CAT}~\cite{haldar2020aMulti} propose a CAT model that uses sequence encoders to represent code snippets based on both raw tokens and the converted string sequence of AST.
The authors also introduce multi-perspective matching operations~\cite{ijcai2017-579} into CAT and build a combined model call MPCAT.
However, we argue that MPCAT is very time-consuming and unrealistic for code retrieval tasks with datasets like ours.
The reason is that our training dataset is roughly \textbf{100 times} the size of the dataset in~\cite{haldar2020aMulti}.
If we train MPCAT with 10 epochs (\ie, the same as our experiment settings for a fair comparison) upon our dataset, it is expected to take about 660 hours (27 days), which is unacceptable in our evaluation.
Thus, we only consider CAT as the baseline model in our evaluation.

\end{itemize}

\subsection{Experimental Settings}\label{sub_sec:experimental_settings}
\subsubsection{Parameter Settings}
To set up our model, we use one layer of RGCN with the output node dimension of 100 and use ReLU as the activation function.
Since each node in the graph contains one word token, we initialize each node with the pre-trained embeddings from GloVe~\cite{pennington2014glove} and the dimension of one word embedding is 300.
For those tokens that cannot be initialized from GloVe, we try to split them into sub-tokens (\eg, CamelCase) and use the average of GloVe pre-trained embeddings of sub-tokens for initialization.
Otherwise, we initialize them with all zeros.
We set the output size of FCMax to 100.

Our implementation is built using the PyTorch~\cite{paszke2019pytorch} and PyTorch\_Geometric~\cite{Fey2019PytorchGeometric} toolkits.
To train our model, we fix the margin $\delta$ to 0.5, set the batch size to 10, and use Adam optimizer with a learning rate of 0.0001.
We train our model for 10 epochs and select the best model based on the lowest validation loss.
In general, it takes approximately 2.5 hours to train, validate and test our model.
Note that, all experiments are conducted on a PC equipped with 8 Intel Xeon 2.2GHz CPU, 256GB memory, and one NVIDIA GTX 1080Ti GPU.

\subsubsection{Evaluation Metrics}\label{subsubsec:evaluationmetric}
In our evaluation, we consider the document description of code as the query text and the code itself as the ground-truth result of code retrieval tasks, which is similar to~\cite{haldar2020aMulti} but different from~\cite{gu2018deep,li2019neural}.
Since the evaluation queries in~\cite{gu2018deep} and~\cite{li2019neural} do not have the ground-truth code snippets,
they use manually labeled ground-truth or utilize another code-to-code similarity tool for judgment.
We argue that both these evaluation methods introduce human bias and manual threshold settings.

Specifically, for each pair of code snippets and descriptions in the testing dataset, we treat the description as the query and take the corresponding code snippet together with the other 99 randomly selected code snippets as the searching candidates for code retrieval tasks.
We adopt two kinds of evaluation metrics that are commonly used in information retrieval to measure the performance of our models and baseline models - mean reciprocal rank (MRR) and success at k (S@k).
Specifically, MRR is the average of the reciprocal ranks of results for a set of queries that denoted as $Q$,
\begin{equation}
  \operatorname*{MRR} = \frac{1}{|Q|} \sum_{q=1}^{|Q|} \frac{1}{\operatorname*{FRank_q}}  
\end{equation}
where $\operatorname*{FRank_q}$ refers to the rank position of the first result for the $q$-th query, $|Q|$ is the number of queries in $Q$.
In addition, S@k denotes the percentage of queries for which more than one correct result exists in the top $k$ ranked results.
\begin{equation}
\text{S@k} = \frac{1}{|Q|} \sum_{q=1}^{|Q|} \Gamma(\operatorname*{FRank_q} \leq k)
\end{equation}
where $\Gamma$ returns 1 if the input is true and returns 0 otherwise.
For both evaluation metrics, a higher metric value implies better performance.

It is noted that for the most complex final model DGMS with SubMul on both datasets, it averagely takes about \textbf{3 hours} for training, validation, and testing in our experiments.

\subsection{Experimental Results \& Discussion}\label{sub_sec:experimental_results}
In this subsection, we empirically compare the proposed model DGMS with 7 state-of-the-art baseline methods and illustrate the effectiveness and generality of the proposed DGMS model.
We also systematically investigate the impact and contribution of each part (\ie, different graph encoding models rather than RGCN, different feature dimensions of RGCN, different semantic matching operations, and different aggregation operations) in the proposed DGMS model by using ablation studies.

\subsubsection{Comparison of Baseline Methods}\label{sub_sub_sec:experiment_baselines}
To demonstrate the effectiveness of the proposed DGMS model, we quantitatively evaluate the DGMS model by measuring and comparing MRR and S@k performance on the testing samples of both \textit{FB-Java} and \textit{CSN-Python} datasets.
Table~\ref{tab:result:main} presents the experimental results of our final model DGMS compared against 7 baseline methods on both datasets.
It is noted that DGMS refers to the final model that applies the SubMul matching operation in the semantic matching module.

Among the 7 baseline methods, it is clearly observed from Table~\ref{tab:result:main} that UNIF shows the best performance on the \textit{FB-Java} dataset, while CAT performs the best on \textit{CSN-Python}.
Interestingly, CAT performs the worst on \textit{FB-Java}, which implies that different baselines methods tend to have different performance on different datasets and some baselines might bias the performance for some datasets.
On the other hand, 1D-CNN performs very poorly on both datasets in terms of all evaluation metrics.
One possible reason we conjecture is that 1D-CNN cannot capture the semantic information for both query texts and code snippets with rich structural and dependency features (\eg, long-range dependencies may exist in code snippets).
Therefore, we suggest that subsequent work should not use 1D-CNN as the encoding model, especially for code snippets.

Compared with all 7 baseline methods, our proposed DGMS model achieves state-of-the-art performance on both datasets in terms of all the four evaluation metrics, \ie, MRR, S@1, S@5, and S@10.
For both \textit{FB-Java} and \textit{CSN-Python} datasets, our model has the performance of over 85\% MRR, over 80\% S@1, and over 95\% S@5, which implies low inspection effort of DGMS to retrieve the desired result for code retrieval tasks.
In particular, for the \textit{CSN-Python} dataset, the DGMS model has significantly higher performance than the best results of the other 7 baseline methods by a large margin up to 22.1, 27.9, 13.9, and 8.6 absolute value on MRR, S@1, S@5, and S@10, respectively.
These correspond to increase rates of performance in MRR, S@1, S@5, and S@10 are high as 31.5\%,  46.7\%, 16.6\%, and 9.5\%, respectively.

S@k is an important evaluation metric for code retrieval tasks.
A higher S@k value implies more likely the correct results exist in the top $k$ ranked returned results.
From Table~\ref{tab:result:main}, we observe that the S@1 score of our DGMS model is 81.7\% for \textit{FB-Java} and 87.6\% for \textit{CSN-Python}.
For both datasets, the S@5 scores of DGMS are over 95\%.
These observations suggest that our DGMS model is more likely (\ie, over 80\% statistically) to get the correct code snippet from the top 1 returned ranked results.
Otherwise, DGMS can easily get the correct code snippet from the top 5 returned ranked results with over 95\% probability statistically.
\begin{table}
  \caption{Experimental results compared with 7 baseline methods.}%
  \centering
    \begin{tabular}{llcccc}
    \toprule
    \multirow{2}{*}{\textbf{Dataset}} & \multirow{2}{*}{Model} & \multicolumn{4}{c}{Result (\%)} \\
                            &       & MRR   & S@1   & S@5   & S@10 \\
    \midrule
    \multirow{8}{*}{\textit{FB-Java}}  & Neural BoW & 77.7 & 71.3 & 85.3 & 88.5 \\
          & RNN                     & 71.7  & 63.0  & 83.2 & 88.6 \\
          & 1D-CNN                  & 22.6  & 12.3  & 32.7 & 45.7 \\
          & Self-Attention          & 65.3  & 54.4  & 79.1 & 84.2 \\
          & DeepCS                  & 78.9  & 70.6  & 89.6 & 94.2 \\
          & UNIF                    & 84.8  & 78.1  & 92.5 & 95.7 \\
          & CAT                     & 20.6  & 10.1  & 28.9 & 41.5 \\
          \cmidrule{2-6}
          & \bf DGMS                & \bf 87.9  & \bf 81.7  & \bf 95.5 & \bf 96.7 \\
    \midrule
    \multirow{8}{*}{\textit{CSN-Python}} & Neural BoW & 66.0 & 56.2 & 78.3 & 83.2 \\
          & RNN                     & 62.7 & 52.8 & 73.1 & 81.6 \\
          & 1D-CNN                  & 18.4 & 10.5 & 25.1 & 33.6 \\
          & Self-Attention          & 63.9 & 54.5 & 75.3 & 82.1 \\
          & DeepCS                  & 64.4 & 52.2 & 78.2 & 88.3 \\
          & UNIF                    & 59.3 & 47.0 & 73.5 & 83.2 \\
          & CAT                     & 70.1 & 59.7 & 83.8 & 90.3 \\
          \cmidrule{2-6}
          & \bf DGMS                & \bf 92.2 & \bf 87.6   & \bf 97.7 & \bf 98.9 \\
    \bottomrule
    \end{tabular}
    \label{tab:result:main}
\end{table}%

\subsubsection{Impact of different graph encoding models.}\label{sub_sub_sec:experiment_diff_graph_encoder}
We investigate the impact of different relational GNNs employed in the graph encoding module of DGMS.
Specifically, we replace RGCN with two variants: Message Passing Network Network (MPNN)~\cite{gilmer2017neural} and Crystal Graph Convolutional Neural (CGCN)~\cite{xie2018crystal}, as both of them can encode graphs with different types of edges.
It is noted that we do not fine-tune any hyper-parameters of the three models and all other experimental settings (\eg, the output dimension of graph encoding, SubMul matching, and FCMax aggregation operations as well as training hyper-parameters) of DGMS (MPNN) and DGMS (CGCN) models are kept the same with DGMS as the previous evaluation.
\begin{table}
  \centering
  \caption{Impact of different graph encoders.}
    \begin{tabular}{lccccc}
    \toprule
    \multirow{2}{*}{\bf Dataset} & \multirow{2}{*}{Model} & \multicolumn{4}{c}{Result (\%)} \\
          &       & MRR   & S@1   & S@5   & S@10 \\
    \midrule
    \multirow{3}{*}{\textit{FB-Java}}   & DGMS (MPNN)    & 85.8 & 79.0 & 94.3 & \bf 96.7 \\
                                                    & DGMS (CGCN)    & 85.6 & 78.6 & 94.4 & 96.0 \\
                                                    & \bf DGMS           & \bf 87.9 & \bf 81.7 & \bf 95.5 & \bf 96.7 \\
    \midrule
    \multirow{3}{*}{\textit{CSN-Python}} & DGMS (MPNN)   & 91.8 & 87.3 & \bf 97.9 & 98.9 \\
                                & DGMS (CGCN)   & \bf 92.5 & \bf 88.2 & \bf 97.9 & \bf 99.1 \\
                                & \bf DGMS          & 92.2 & 87.6 & 97.7 & 98.9 \\
    \bottomrule
    \end{tabular}\label{tab:result:differentgnn}%
\end{table}%

Table~\ref{tab:result:differentgnn} presents the results of RGCN versus MPNN/CGCN in the DGSM architecture.
We can observe that all three models achieve similar and stable performance on both datasets, implying our model architecture is not sensitive to different relational GNNs in the graph encoding module. 
Interestingly, DGMS (CGCN) slightly improves DGMS on the \textit{CSN-Python} dataset, which indicates that our model can be further improved by carefully choosing other graph encoding models according to different application tasks.

\subsubsection{Impact of different feature dimensions of RGCN}\label{sub_sub_sec:experiment_diff_feat_dims}
\begin{figure*}[t]
\centering
\subfigure[MRR]{\includegraphics[width=0.24\linewidth]{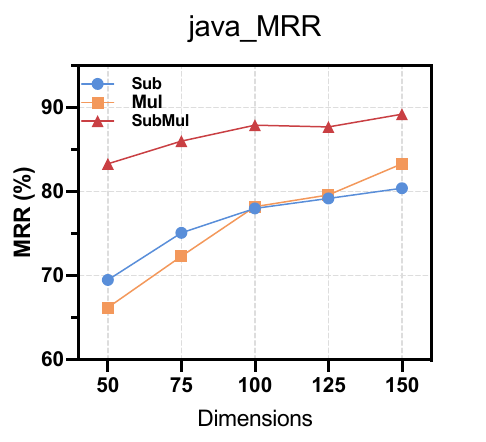}}
\hfill
\subfigure[S@1]{\includegraphics[width=0.24\linewidth]{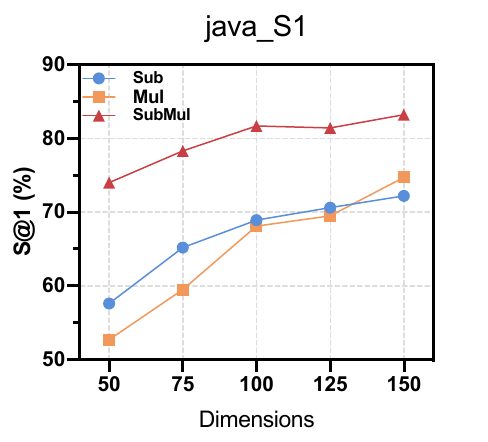}}
\hfill
\subfigure[S@5]{\includegraphics[width=0.24\linewidth]{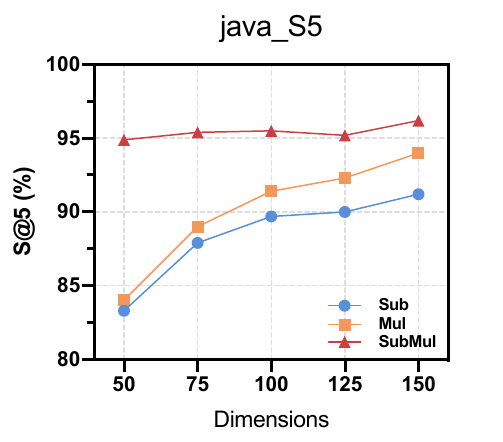}}
\hfill
\subfigure[S@10]{\includegraphics[width=0.24\linewidth]{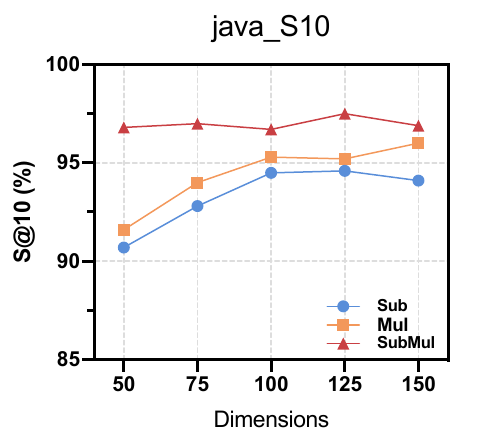}}
\caption{The impact of different feature dimensions in RGCN for the \textit{FB-Java} dataset with regard to MRR, S@1, S@5, and S10.}
\label{fig:ablation:differentfilter_java}
\end{figure*}
\begin{figure*}
\centering
\subfigure[MRR]{\includegraphics[width=0.24\linewidth]{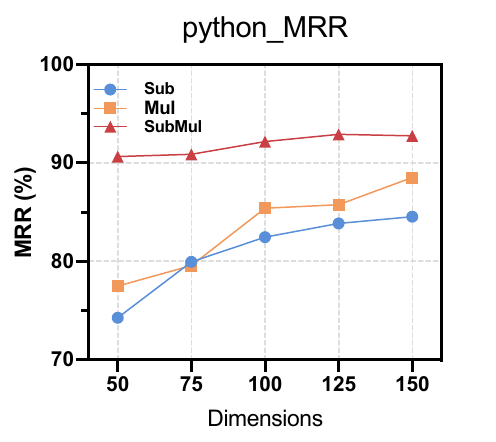}}
\hfill
\subfigure[S@1]{\includegraphics[width=0.24\linewidth]{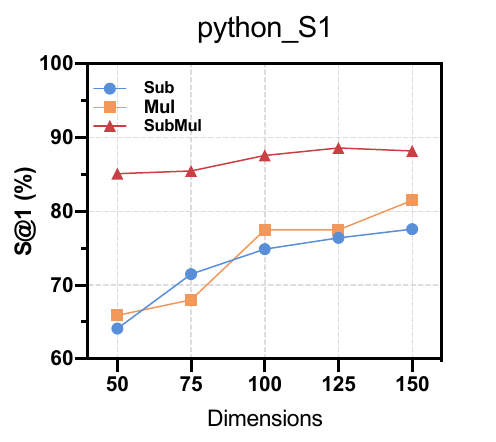}}
\hfill
\subfigure[S@5]{\includegraphics[width=0.24\linewidth]{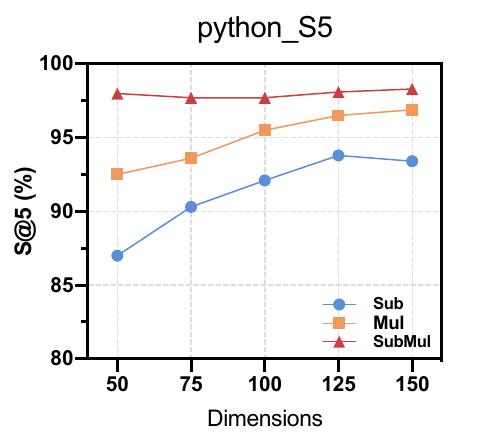}}
\hfill
\subfigure[S@10]{\includegraphics[width=0.24\linewidth]{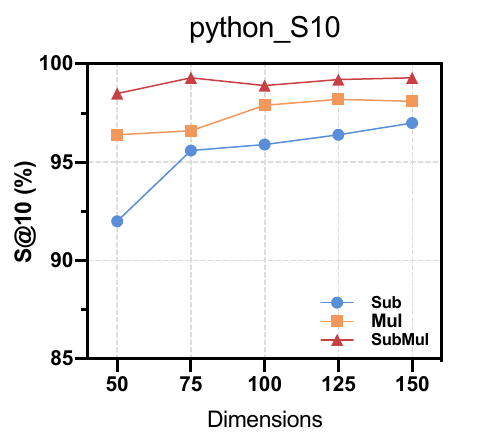}}
\caption{The impact of different feature dimensions in RGCN for the \textit{CSN-Python} dataset with regard to MRR, S@1, S@5, and S10.}
\label{fig:ablation:differentfilter_python}
\end{figure*}

We further study the impact of different feature dimensions in graph encoder (\ie, RGCN) on the performance of our DGMS models.
Following the default and same parameter settings of previous experiments, we only change the output dimension of RGCN ($d=50/75/100/125/150$) and observe their impacts on the performance of DGMS-Sub, DGMS-Mul, and DGMS-SubMul. 
The main results for both \textit{FB-Java} and \textit{CSN-Python} datasets are shown in Figure~\ref{fig:ablation:differentfilter_java} and Figure~\ref{fig:ablation:differentfilter_python}, respectively.

It is clearly seen that the performance (\ie, MRR, S@1, S@5, and S@10) of DGMS-Sub, DGMS-Mul, and DGMS-SubMul improves as the feature dimension in the graph encoder grows.
The reason is intuitive that increasing feature dimensions in the graph encoder imply increasing model capacities, which usually show better performance in supervised learning.
For DGMS-Sub and DGMS-Mul, their performance increases rapidly with the increasing feature dimensions,
while our final DGMS-SubMul model shows a relatively slower increasing trend.
We conjecture this is because the model capacity of DGMS-SubMul is comparably more than DGMS-Sub and DGMS-Mul.
However, it is a trade-off between model performance and time/memory consumption since the increasing features dimensions of the three models inevitably bring more time and memory consumption.
We thus set the default feature dimension in RGCN of all three models to be 100 for other evaluations.

\subsubsection{Impact of different semantic matching operations.}\label{sub_sub_sec:experiment_diff_matching}
\begin{table}[htb]
   \centering
    \caption{Impact of different matching operations.}
    \begin{tabular}{llcccc}
    \toprule
    \multirow{2}{*}{\bf Dataset} & \multirow{2}{*}{Model} & \multicolumn{4}{c}{Result (\%)} \\
          &       & MRR   & S@1   & S@5   & S@10 \\
    \midrule
    \multirow{4}{*}{\textit{FB-Java}} 
          & DGMS-No     & 75.5 & 64.3 & 89.6 & 93.8 \\
          & DGMS-Sub    & 78.0 & 68.9 & 89.7 & 94.5 \\
          & DGMS-Mul    & 78.2 & 68.1 & 91.4 & 95.3 \\
          & \bf DGMS    & \bf 87.9 & \bf 81.7 & \bf 95.5 & \bf 96.7 \\
    \midrule
    \multirow{4}{*}{\textit{CSN-Python}}
          & DGMS-No     & 79.2 & 70.7 & 90.5 & 94.8 \\
          & DGMS-Sub    & 82.5 & 74.9 & 92.1 & 95.9 \\
          & DGMS-Mul    & 85.4 & 77.5 & 95.5 & 97.9 \\
          & \bf DGMS    & \bf 92.2 & \bf 87.6 & \bf 97.7 & \bf 98.9 \\
    \bottomrule
    \end{tabular}
    \label{tab:result:ablation}
\end{table}%
In order to assess the impact of different operations of the semantic matching module in our model, we conduct ablation studies to evaluate the performance of the following model variants with different semantic matching operations:
\begin{itemize}
    \item DGMS-No refers to the model that does not employ any semantic matching operation and we apply FCMax directly after the graph encoding;
    \item DGMS-Sub refers to the model with the Sub matching operation as Equation~(\ref{equation:sub});
    \item DGMS-Mul is a variant model with the Mul matching operation as Equation~(\ref{equation:mul});
    \item DGMS is the final model using the SubMul matching operation as Equation~(\ref{equation:submul}).
\end{itemize}
As shown in Table~\ref{tab:result:ablation}, models that employ matching operations (\ie, Sub, Mul, SubMul) achieve noticeably better performance than that without the matching operation (DGMS-No).
These observations highlight the importance of the semantic matching module, which could significantly improve the effectiveness of our DGMS model.
On the other hand, compared with both DGMS-Sub and DGMS-Mul models, the final DGMS model (DGMS-SubMul) achieves better performance, indicating the concatenation of Sub and Mul matching operations captures more interaction features than each of them individually.

Interestingly, compared with all 7 baselines in Table~\ref{tab:result:main}, even DGMS-No performs better than most baselines, especially on the \textit{CSN-Python} dataset.
This observation indicates the usefulness of the graph-structured data and the encoder for representing both the query text and code snippets in our model, as the graph representation learning captures more semantic information. 

\subsubsection{Impact of different aggregation operations}\label{sub_sub_sec:experiment_diff_aggregation}
\begin{table}[htb]
  \centering
  \caption{The impact of different aggregation operations on both datasets.}
    \begin{tabular}{lllcccc}
    \toprule
    {\bfseries Datasets} & Models & Aggregations & MRR   & S@1   & S@5   & S@10 \\
    \midrule
    \multirow{12}{*}{\textit{FB-Java}} 
          & \multirow{4}{*}{DGMS-Sub} & Average & 64.4  & 51.7  & 79.7  & 89.7 \\
          &       & Max   & 66.6  & 53.9  & 83.0  & 90.4 \\
          &       & FCAvg & 73.7  & 62.3  & 87.5  & 92.6 \\
          &       & FCMax & \bf 78.0  & \bf 68.9  & \bf 89.7  & \bf 94.5 \\
    \cmidrule{2-7}          
          & \multirow{4}{*}{DGMS-Mul} & Average & 45.8  & 28.3  & 70.0  & 84.5 \\
          &       & Max   & 51.5  & 33.6  & 75.2  & 87.6 \\
          &       & FCAvg & 78.1  & \bf 68.2  & 91.0  & 95.1 \\
          &       & FCMax & \bf 78.2  & 68.1  & \bf 91.4  & \bf 95.3 \\
    \cmidrule{2-7}          
          & \multirow{4}{*}{DGMS-SubMul} & Average & 83.8  & 74.9  & \bf 95.9  & 96.8 \\
          &       & Max   & 81.8  & 71.3  & 95.5  & 96.0 \\
          &       & FCAvg & 85.6  & 78.1  & 94.8  & \bf 97.0 \\
          &       & FCMax & \bf 87.9  & \bf 81.7  & 95.5  & 96.7 \\
    \midrule
    \multirow{12}{*}{\textit{CSN-Python}}
          & \multirow{4}{*}{DGMS-Sub} & Average & 62.4  & 49.2  & 79.5  & 89.3 \\
          &       & Max   & 65.6  & 53.5  & 82.4  & 90.7 \\
          &       & FCAvg & 74.9  & 64.1  & 88.5  & 94.8 \\
          &       & FCMax & \bf 82.5  & \bf 74.9  & \bf 92.1  & \bf 95.9 \\
    \cmidrule{2-7}          
          & \multirow{4}{*}{DGMS-Mul} & Average & 46.7  & 27.7  & 71.9  & 86.6 \\
          &       & Max   & 48.0  & 28.9  & 73.7  & 88.0 \\
          &       & FCAvg & 81.6  & 72.4  & 92.4  & 96.6 \\
          &       & FCMax & \bf 85.4  & \bf 77.5  & \bf 95.5  & \bf 97.9 \\
    \cmidrule{2-7}
          & \multirow{4}{*}{DGMS-SubMul} & Average & 89.5  & 83.1  & 97.2  & 98.7 \\
          &       & Max   & 85.7  & 76.7  & 97.0  & 98.7 \\
          &       & FCAvg & 89.7  & 84.0  & 96.8  & 98.4 \\
          &       & FCMax & \bf 92.2  & \bf 87.6  & \bf 97.7  & \bf 98.9 \\
    \bottomrule
    \end{tabular}%
  \label{tab:ablation:different_agg}%
 \end{table}%
We further investigate the impact of different aggregation operations on the performance of our final DGMS models.
Keeping all other experimental settings the same as previous experiments, we employ four different aggregation operations to aggregate a graph-level embedding from the learned node embeddings of the code graph or the text graph.
In general, an aggregation operation must operate over an unordered set of vectors and be invariant to permutations of its inputs.
Average and Max are two simple and straightforward pooling operations that take the element-wise average operation and the element-wise max operation of the input node embeddings, respectively.
FCAvg and FCMax are two variants of Average pooling and Max pooling operations following a fully connected layer transformation.
The main results are summarized in Table~\ref{tab:ablation:different_agg}.

Obviously, it can be seen that all three models (\ie, DGMS-Sub, DGMS-Mul, and DGMS-SubMul) with both FCAvg and FCMax aggregation operations demonstrate superior performance on both datasets in terms of all evaluation metrics.
This implies that the fully connected layer transformation after the max/average pooling plays an important role in aggregating the graph-level embedding.
In addition, FCMax shows better performance than FCAvg for all three models on both \textit{FB-Java} and \textit{CSN-Python} datasets.
We conjecture this is because the max pooling operation could easily compare and find the similarity and dissimilarity of nodes of the text and code graphs, while the average pooling operation might ``flatten'' the aggregated graph-level embedding from node embeddings.

%% file: 4RelatedWork.tex
\section{Related Work}\label{sec:related_work}
\subsection{Deep Learning for Code Retrieval Tasks}
Recently, with the enormous success achieved in many different fields, deep learning techniques have been gradually studied to improve the performance of source code retrieval techniques~\cite{gu2018deep,sachdev2018retrieval,cambronero2019deep,yao2019coacor,wan2019multi,husain2019codesearchnet,haldar2020aMulti}.
\cite{gu2018deep} is the first work that applies deep neural networks for the code retrieval task.
It first simply encodes both the query text and code snippet into a vector representation using MLP or RNN and then try to rank these code snippets n the candidate set according to the similarity score between the learned representations.
Other following work is similar to~\cite{gu2018deep} with only a slight difference in choosing the encoding models.
For instance, {Cambronero}~\etal~\cite{cambronero2019deep} utilized FastText~\cite{bojanowski2017enriching} to initialize the embeddings of all the tokens in queries and codes, and then aggregated them with learnable attention weight or simply averages them;
{Husain}~\etal~\cite{husain2019codesearchnet} learned the tokens embeddings using different standard sequence models (\eg, Neural BoW, RNN, 1D-CNN, and BERT).
{Yao}~\etal~\cite{yao2019coacor} explored a novel perspective of generating code annotations for code retrieval based on the reinforcement learning framework.
It first trains a code annotation model via reinforcement learning to generate a natural language annotation and then uses the generated annotation to better distinguish relevant code snippets from others.
{Haldar}~\etal~~\cite{haldar2020aMulti} used the LSTM encoders to represent code snippets based on both raw tokens and the converted string sequence of AST and explored a bilateral multi-perspective matching model for semantic code searching.

All these work share a similar spirit that first maps both code and natural language description into vectors in the same embedding space with sequence encoders (\eg, RNN, LSTM, Attention, etc.), and then computes the cosine or L2 similarity of these vectors.
However, DGMS differs from previous work in two major dimensions:
1) we first propose to represent both code snippets and query texts with the unified graph-structured data, which largely preserves their structural and semantic information and can also be learned with graph neural networks;
2) DGMS not only captures the semantic information for individual code snippets or query texts, but also explore more fine-grained semantic relation between them for better representation.

\subsection{Other Source Code Related Tasks}
Other active research areas that involve machine learning in source code related tasks 
include code summarization~\cite{iyer2016summarizing,fernandes2019structure,alon2019code2seq,alon2019code2vec,leclair2020improved,ahmad2020transformer}, code generation~\cite{oda2015learning,brockschmidt2019generative}, etc.
For code summarization,
{Iyer}~\etal~\cite{iyer2016summarizing} proposed the first completely data-driven approach for generating high-level summaries of code snippets by employing LSTM with the attention mechanism.
{Fernandes}~\etal~\cite{fernandes2019structure} presented a hybrid model that extends standard sequence encoder models with graph neural networks that leverage additional structure in sequence data to summarize source codes.
{Alon}~\etal~\cite{alon2019code2seq} presented a novel code-to-sequence (code2seq) model which samples paths in the abstract syntax tree of code snippets, encodes these paths with LSTM networks, and attends to them while generating the target sequence.
{Alon}~\etal~\cite{alon2019code2vec} also proposed a new attention-based neural network - code2vec - for representing arbitrary-sized source codes using a learned fix-length continuous vector.
The code2vec uses the soft-attention mechanism over syntactic paths of the abstract syntax tree of the source code and aggregates all of their vector representation into a contextual vector.
{Ahmad}~\etal~\cite{ahmad2020transformer} explored the Transformer~\cite{vaswani2017attention} model that uses the self-attention mechanism to capture the long-range dependencies in code tokens. LeClair ~\etal~\cite{leclair2020improved} also exploited improving code summarization via a graph neural network. 
For code generation,
{Oda}~\etal~\cite{oda2015learning} leveraged the statistical machine translation framework to automatically generate pseudo-code.
{Brockschmidt}~\etal~\cite{brockschmidt2019generative} proposed a novel generative model that uses graphs to represent intermediate states of the generated output.
It generates the code snippet by interleaving grammar-driven expansion steps with graph augmentation and neural message passing steps.
{Alon}~\etal~\cite{alon2020structural} presented a novel structural language model that estimates the probability of AST of code snippet by decomposing it into a product of conditional probabilities over the nodes.

%% file: 5Conclusion.tex
\section{Conclusion}\label{sec:conclusion}
We propose DGMS, a deep graph matching and searching model for semantic code retrieval tasks.
In particular, we represent both source codes and query texts with graph-structured data and then encode and match them with the proposed DGMS model.
Our model makes better use of the rich structural information in source codes and query texts as well as the interaction semantic relations between each other.
Extensive experiments demonstrate that DGMS significantly outperforms the state-of-the-art baseline models by a large margin on two benchmark datasets from two representative programming languages (\ie, Java and Python).
One limitation of our work is that we consider the document descriptions of code snippets as the query texts to train our model as there is no benchmark dataset that contains a large corpus of paired human-written questions and corresponding code snippets.
In future work, we hope to find more datasets with large amounts of pairs of human-written query questions and corresponding code snippets, and then train and test our model with these datasets from a more realistic scenario.
Another future work is to explore more effective ways to construct text graphs and code graphs for other programming languages (\ie, C/C++, C\#, PHP, etc.) as well as develop better graph comparison functions to further improve the performance of semantic code retrieval tasks.